\documentclass[10pt,twocolumn,letterpaper]{article}

\usepackage[pagenumbers]{wacv} 

\usepackage{graphicx}
\usepackage{amsmath}
\usepackage{amssymb}
\usepackage{booktabs}

\usepackage{multirow}
\usepackage{algorithm}
\usepackage{algorithmic}
\usepackage{multirow}
\usepackage{xcolor}
\usepackage{amsfonts}
\usepackage{pifont}
\usepackage{tabularx}
\newcommand{\cmark}{\ding{51}} 
\newcommand{\xmark}{\ding{55}} 

\usepackage[pagebackref,breaklinks,colorlinks]{hyperref}

\usepackage[capitalize]{cleveref}
\crefname{section}{Sec.}{Secs.}
\Crefname{section}{Section}{Sections}
\Crefname{table}{Table}{Tables}
\crefname{table}{Tab.}{Tabs.}

\begin{document}

\title{AutoProSAM: Automated Prompting SAM for 3D Multi-Organ Segmentation}

\author{
Chengyin Li$^{1}$ \quad Rafi Ibn Sultan$^{1}$ \quad Prashant Khanduri$^{1}$ \quad \\ Yao Qiang $^{1}$ \quad Chetty Indrin$^{2}$ \quad Dongxiao Zhu$^{1}$\thanks{Corresponding author.}\\ 
 $^{1}$Wayne State University \quad $^{2}$Cedars Sinai Medical Center\\
{\tt \small \{cyli, rafisultan, khanduri.prashant, yao, dzhu\}@wayne.edu \quad indrin.chetty@cshs.org}
}

\maketitle

\begin{abstract}
Segment Anything Model (SAM) is one of the pioneering prompt-based foundation models for image segmentation and has been rapidly adopted for various medical imaging applications. However, in clinical settings, creating effective prompts is notably challenging and time-consuming, requiring the expertise of domain specialists such as physicians. This requirement significantly diminishes SAM's primary advantage—its interactive capability with end users—in medical applications. Moreover, recent studies have indicated that SAM, originally designed for 2D natural images, performs suboptimally on 3D medical image segmentation tasks. This subpar performance is attributed to the domain gaps between natural and medical images and the disparities in spatial arrangements between 2D and 3D images, particularly in multi-organ segmentation applications. To overcome these challenges, we present a novel technique termed \textbf{AutoProSAM}. This method automates 3D multi-organ CT-based segmentation by leveraging SAM's foundational model capabilities without relying on domain experts for prompts. The approach utilizes parameter-efficient adaptation techniques to adapt SAM for 3D medical imagery and incorporates an effective automatic prompt learning paradigm specific to this domain. By eliminating the need for manual prompts, it enhances SAM's capabilities for 3D medical image segmentation and achieves state-of-the-art (SOTA) performance in CT-based multi-organ segmentation tasks. The code is in this \href{https://github.com/ChengyinLee/AutoProSAM_2024}{link}.

\end{abstract}

\section{Introduction}
\label{sec:intro}

Recently, prompt-based foundation models in computer vision, such as the Segment Anything Model (SAM) ~\cite{kirillov2023segment}, has shown impressive performance and versatility across various semantic segmentation tasks~\cite{zhang2023comprehensive} based on user-provided prompts. These foundation models offer new potential for multi-organ segmentation in medical imagery, an area often limited by the availability and quality of segmentation masks. Unlike custom-designed transformer models such as UNETR~\cite{hatamizadeh2022unetr}, SwinUNETR~\cite{tang2022self}, and FocalUNETR~\cite{li2023focalunetr}, which are typically trained on limited patient samples and masks, foundation models like SAM benefit from training on millions of 2D natural images and billions of masks. With appropriate prompts, these foundation models can be adapted to a variety of downstream tasks.

\begin{figure}[t]
\centering
\includegraphics[width=0.97\columnwidth]{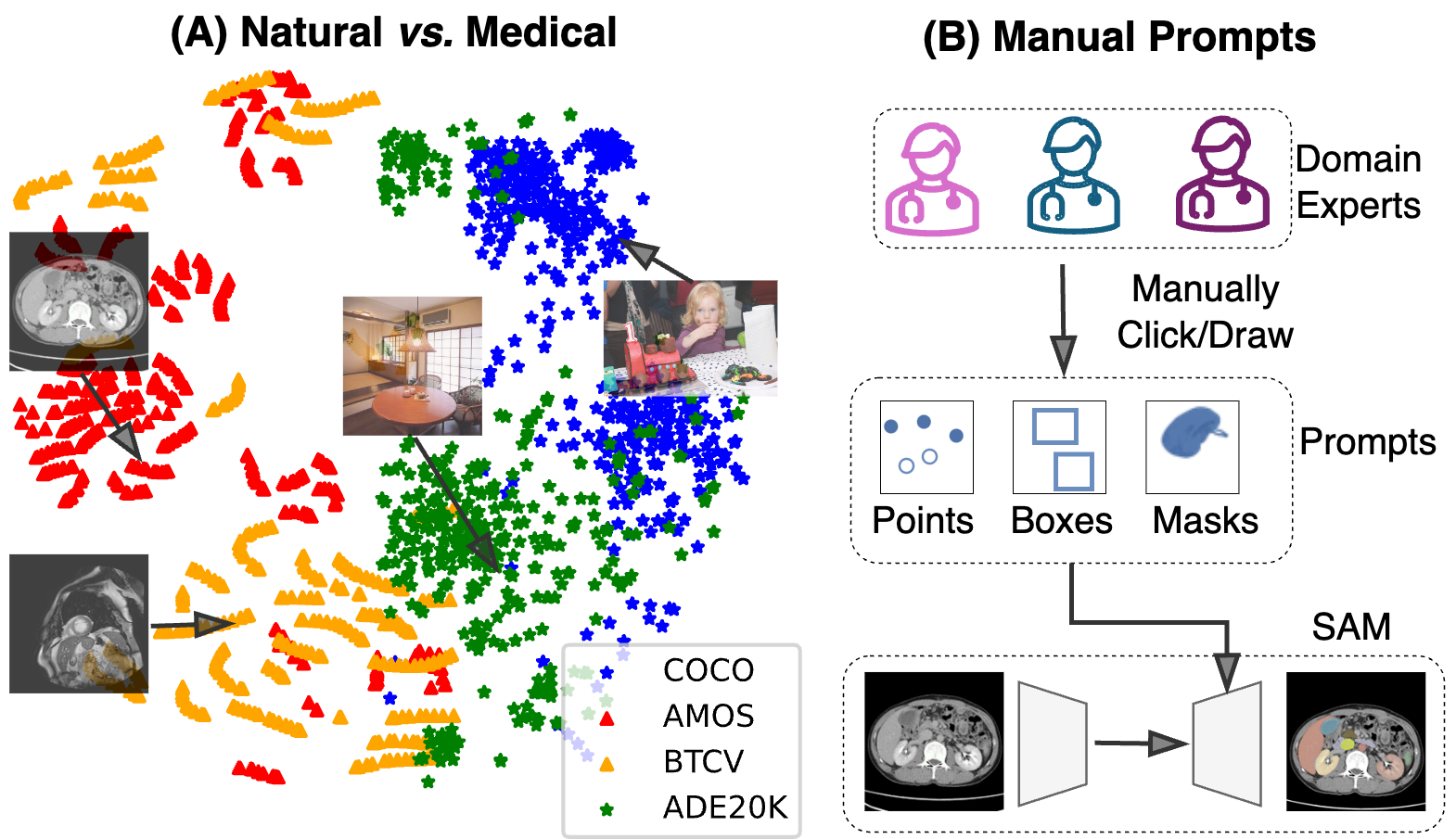} 
\caption{Challenges associated with using SAM for medical image segmentation include (A) a T-SNE plot of embeddings encoded by SAM's image encoder, showcasing differences between medical image datasets such as AMOS~\cite{ji2022amos} and BTCV~\cite{landman2015miccai}, and natural image datasets like ADE20K~\cite{zhou2017scene} and COCO~\cite{lin2014microsoft}; (B) The requirement of manually generated prompts from domain experts for SAM-based medical image segmentation.}
\label{fig1}
\end{figure}

Despite their extensive training, large foundation models such as SAM struggle to adapt to medical images. This is primarily due to the inherent differences between medical and natural images and the challenges associated with effective prompting (as shown in \Cref{fig1}). Recent efforts have attempted to extend the success of SAM to medical image segmentation tasks~\cite{SAM4MIS,wu2023medical,ma2023segment,gong20233dsam,shaharabany2023autosam}. However, the demonstrated performance has exhibited reduced precision and stability, particularly in more intricate segmentation tasks in 3D medical images characterized by smaller sizes, irregular shapes, and lower contrast properties compared to 2D natural images~\cite{gong20233dsam}. \Cref{fig1}(A) illustrates this oversight of substantial domain disparities between natural and medical images. Furthermore, unlike natural images, medical images in various formats are not easily interpretable by general users. The prompts, intended to guide the foundation models to focus on a particular location, become a performance bottleneck as users struggle to identify the region of interest. Therefore, generating effective prompts for these models necessitates the involvement of domain experts and physicians. This reliance on labor-intensive manually generated prompts~\cite{gao2023desam,shaharabany2023autosam} hampers its successful application, particularly in multi-organ medical image segmentation tasks, as demonstrated in \Cref{fig1}(B). As a result, these foundation models have been shown to underperform in medical image segmentation tasks compared to state-of-the-art (SOTA) models specifically designed for medical imagery~\cite{gong20233dsam}.

To address the issues mentioned above, we introduce a novel \textbf{AutoProSAM} method for transitioning SAM from 2D natural image segmentation to 3D multi-organ image segmentation without relying on human prompts. We start by designing extensive modifications for the image encoder at the input level, allowing the original 2D transformer to handle volumetric 3D inputs effectively. This optimization enhances the reusability of pre-trained weights through a parameter-efficient fine-tuning method, achieving a balance by leveraging SAM's vast knowledge from training on a large 2D natural image dataset and adapting it with minimal tuning for 3D medical images. To solve the prompting issue and adapt SAM for end users, we developed an Auto Prompt Generator (APG) module that automatically learns the necessary prompts for the segmentation process. This innovation eliminates the time-consuming and complex manual prompt generation process, particularly for multi-organ medical image segmentation tasks. Our extensive experimentation on CT-based multi-organ segmentation datasets, which includes comprehensive comparisons with state-of-the-art methods such as nnUNet~\cite{isensee2021nnu} and recent adapters, shows a significant performance improvement over existing techniques.

The main contributions are three-fold: 
\begin{itemize}
    \item We introduce the Auto Prompt Generator to eliminate the complex and laborious manual prompting process, simplifying multi-organ medical image segmentation tasks for non-domain-expert users.
    \item We adapt the 2D SAM model for 3D multi-organ segmentation with parameter-efficient fine-tuning, bridging the gap between 2D natural images and 3D medical images.
    \item We conduct extensive experiments and analysis on three public and one private multi-organ segmentation dataset, demonstrating that the proposed \textbf{AutoProSAM} achieves superior performance in medical image segmentation tasks compared to the baselines.
\end{itemize}

\section{Related Work}
\label{sec:1related_work}

\begin{figure*}[t]
\centering
\includegraphics[width=1.0\textwidth]{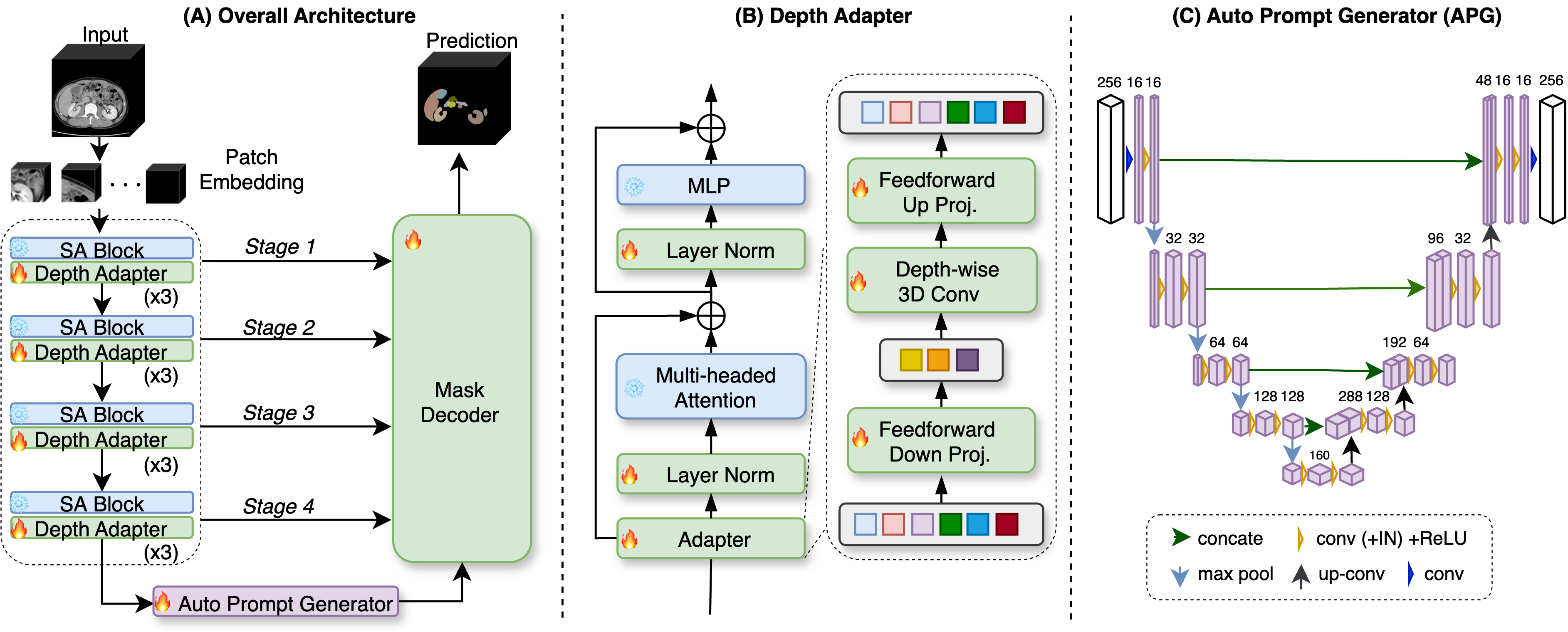} 
\caption{(A) The overall architecture of the AutoProSAM, (B) the design of the Depth Adapter module, which utilizes parameter-efficient model fine-tuning, and (C) the architecture of the Auto Prompt Generator, featuring a U-Net-like encoder-decoder design.}
\label{fig2}
\end{figure*}

\subsection{Foundation Computer Vision Models}
Recent advancements in computer vision have led to the development of numerous pre-trained backbones using various algorithms and datasets~\cite{goldblum2023battle}. As deep learning models evolve, most modern vision frameworks now adhere to the pre-training and fine-tuning paradigm~\cite{min2021recent}. Recently, there has been significant interest among computer vision researchers in large, adaptable foundation models leveraging pre-training techniques such as self-supervised learning~\cite{jing2020self}, contrastive learning~\cite{wang2022contrastive}, and language-vision pre-training~\cite{radford2021learning}. Notably, the SAM model~\cite{kirillov2023segment}, pre-trained on a dataset of over 11 million images, has emerged as a prompt-based foundation model for natural image segmentation. SAM demonstrates impressive zero-shot capabilities, effectively segmenting diverse subjects in real-world environments through an interactive and prompt-driven approach. Additionally, SEEM~\cite{zou2023segment}, developed around the same time as SAM, introduces a more comprehensive prompting scheme to facilitate semantic-aware open-set segmentation. Furthermore, DINOv2~\cite{oquab2023dinov2} focuses on scaling up the pre-training of a Vision Transformer (ViT) model in terms of data and model size, aiming to produce versatile visual features that simplify the fine-tuning of downstream tasks.

\subsection{Parameter-efficient Model Fine-Tuning}
Given the extensive utilization of foundation models, the concept of parameter-efficient fine-tuning has garnered significant attention. Existing methods for efficient fine-tuning can be categorized into three groups~\cite{ding2023parameter}. First, addition-based methods incorporate lightweight adapters~\cite{pan2022st, wang2023med} or prompts~\cite{liu2023pre, jia2022visual} into the original model, focusing solely on adjusting these parameters. Second, specification-based methods~\cite{zaken2021bitfit, guo2020parameter} concentrate on selecting a small subset of the original parameters for tuning. Third, reparameterization-based methods~\cite{hu2021lora} leverage low-rank matrices to approximate parameter updates. Recently, some researchers have extended pre-trained image models to encompass video comprehension~\cite{pan2022st} or volumetric segmentation~\cite{wang2023med}. However, these methods treat the additional dimension as a ``word group" and employ specialized modules to aggregate information along the word dimension. In contrast, our work considers all three dimensions simultaneously and directly adapts the trained transformer block to capture patterns in 3D medical image inputs.

\subsection{Adapting SAM to Medical Images}
Among the efforts to adapt SAM for medical images, MedSAM~\cite{ma2023segment} focused on refining the SAM decoder using prompts generated from label masks across more than 30 medical image datasets, resulting in improved performance over zero-shot predictions. Zhang \etal~\cite{zhang2023customized} optimized a low-rank fine-tuning approach for the SAM encoder and combined it with SAM decoder training for abdominal segmentation tasks. Tal \etal~\cite{shaharabany2023autosam} redesigned SAM's prompt encoder by using the original 2D medical image as input. SAM-Med2D~\cite{cheng2023sam} fine-tuned SAM with a large collection of 2D medical image datasets. Conversely, Wu \etal~\cite{wu2023medical} pre-trained the entire encoder through self-supervised learning on collected datasets for downstream medical image segmentation tasks. 3DSAM-Adapter~\cite{gong20233dsam} extended SAM to 3D medical imagery, targeting tumor segmentation tasks for each tumor type individually, a time- and labor-intensive process due to the demanding prompting requirements. Similarly, Wang \etal~\cite{wang2023sam} addressed 3D medical imagery by retraining SAM from scratch, replacing the 2D components to develop an in-house SAM for medical imagery. Despite some progress, existing methods often overlook the 3D patterns of medical images or rely heavily on extensive pre-training.

\subsection{Automatic Prompts Generation}
To automate the prompt generation process for SAM, research has primarily taken two directions. One direction, exemplified by works such as ~\cite{pandey2023comprehensive,lei2023medlsam}, involves generating 2D bounding box prompts for either 2D image segmentation or each 2D slice in 3D images. While these methods are a good starting point, they do not address the 3D spatial relationships inherent in 3D images. Another direction, seen in efforts such as ~\cite{shaharabany2023autosam,na2024segment,colbert2024repurposing}, involves training an auxiliary network to generate prompts for SAM from the input image. However, these models fail to fully utilize SAM's extensive pre-trained capabilities. Our approach overcomes these limitations by leveraging features from SAM's encoder to generate automated prompts. Consequently, we address the shortcomings of previous works and unlock SAM's full potential in CT-based 3D multi-organ segmentation tasks.

\section{Method}
\label{sec:method}

In this section, we explain how to modify the original SAM architecture, initially developed for 2D natural images, to work with 3D volumetric medical images for segmentation tasks. We begin by offering a brief overview of the SAM framework (as depicted in \Cref{fig2}), followed by a detailed explanation of the adjustments made to the image encoder, prompt encoder, and mask decoder.

\subsection{SAM Architecture}
SAM~\cite{kirillov2023segment} is a prompt-driven image segmentation framework, renowned for its exceptional performance in segmenting natural images. The architecture of SAM comprises three key components: an image encoder, a prompt encoder, and a mask decoder. The image encoder utilizes the Vision Transformer (ViT)~\cite{dosovitskiy2020image,qiang2023interpretability} to transform original images into one-time image embeddings. The prompt encoder skillfully converts various types of prompts – including foreground/background points, rough boxes or masks, clicks, text, or any information indicating the target of segmentation – into compact embeddings. These embeddings from both the image and prompt encoders are then seamlessly integrated by the mask decoder to produce accurate segmentation masks.

Although SAM has been successful in 2D natural image segmentation, it faces significant challenges when applied to 3D volumetric medical imagery. A key issue is the model's reliance on slice-wise predictions, which fail to consider the inter-slice spatial context, thereby impacting its suitability for complex medical tasks. Additionally, the inherent domain disparities between medical and natural images contribute to their performance limitations in medical applications. To effectively address these challenges and optimize SAM for medical imaging tasks, tailored adaptation, and fine-tuning of the model become essential.

\subsection{Handling 3D Medical Inputs}

To enhance SAM's capability in processing 3D medical images, we propose an adaptation strategy named \textbf{AutoProSAM}, as illustrated in \Cref{fig2}A. This strategy has two primary objectives: firstly, to enable the model to directly learn 3D spatial patterns, and secondly, to ensure continuity by inheriting most parameters from the pre-trained model while introducing easily adjustable incremental parameters. The detailed design is elaborated as follows.

\subsubsection{Positional Encoding Enhancement}
In the pre-trained ViT model, there is a lookup table of size $C \times H \times W$ for positional encoding, where $C$ represents the channel, $H$ the height, and $W$ the width. Additionally, we initialize a tunable lookup table of size $C \times D$ (with $D$ representing the depth of a volume patch) with zeros.  To obtain the positional encoding of a 3D point $(d,h,w)$, we add the embedding from the frozen lookup table with $(h,w)$ to the embedding from the tunable lookup table with $(d)$.

\subsubsection{Patch Embedding Adjustments}
We utilize a combination of $1\times k\times k$ and $k \times 1\times 1$ 3D convolutions to approximate the effect of a $k \times k\times k$ convolution (\eg, kernel size $k=14$). The $1\times k\times k$ convolution is initialized with the weights from a pre-trained 2D convolution and remains unchanged during the fine-tuning phase. As for the newly introduced $k \times1 \times 1$ 3D convolution, we apply depth-wise convolution to decrease the number of parameters that need adjustment. This approach helps in managing the complexity of the model. 

\subsubsection{Adapting Attention Block}
The attention blocks can be directly adjusted to accommodate 3D features. In the case of 2D inputs, the size of the queries is $[B, HW, C]$, which can be effortlessly modified to $[B, DHW, C]$ for 3D inputs, while retaining all the pre-trained weights. We implement sliding-window mechanisms akin to those in the SwinUNETR~\cite{tang2022self} to mitigate the memory impact resulting from the increase in dimensions. This approach aids in optimizing the model's performance while managing memory requirements.

\subsubsection{Bottleneck Modifications}
Given that convolution layers are generally easier to optimize than transformers, we replace 2D convolutions in the bottleneck with 3D counterparts and train them from scratch to improve performance.

By making the above adjustments, we can smoothly transition the 2D ViT into a 3D ViT, reusing most parameters. However, fully fine-tuning the 3D ViT can be resource-intensive. To address this, we propose using a lightweight adapter approach for efficient fine-tuning. The adapter comprises a down-projection linear layer and an up-projection linear layer, represented as $\operatorname{Adapter}(\mathbf{X}) = \mathbf{X} + \operatorname{Act}(\mathbf{X} W_{\mathrm{Down}}) W_{\mathrm{Up}}$. Here, $\mathbf{X} \in \mathbb{R}^{N \times C}$ is the original feature representation, $W_{\mathrm{Down}} \in \mathbb{R}^{C\times N'}$ and $W_{\mathrm{Up}} \in\mathbb{R}^{N'\times C}$ are down-projection and up-projection layers, and $\mathrm{Act}(\cdot)$ is the activation function. To enhance 3D spatial awareness, we include a depth-wise 3D convolution after the down-projection layer, as shown in \Cref{fig2}B. This enhancement improves the adapter's utilization of 3D spatial cues.

Throughout the training phase, we exclusively adjust the parameters of convolutions, depth adapters, and normalization layers, while maintaining all other parameters in a frozen state. This frozen approach enhances memory efficiency during training. Fine-tuning the adapter and normalization layers aids in bridging the gap between natural images and medical images, enabling the model to adapt more effectively to the medical image domain.

\subsection{Auto Prompt Generator}
SAM initially applies positional embedding to both prompt (\eg, points or boxes) and image, ensuring that prompt and image embeddings at the same position share identical positional encoding. Subsequently, the prompt embedding engages in cross-attention with the image embedding, evolving from positional to semantic attributes. However, this cross-attention mechanism, while effective in 2D settings, can lead to over-smoothing issues when applied to 3D feature maps~\cite{gong20233dsam}. Adapting to 3D settings can significantly increase the number of tokens, potentially resulting in a uniform probability distribution.

Furthermore, prompt-based segmentation may not be well-suited for real-world applications due to two primary reasons. Firstly, it is time-consuming for multi-class scenarios, as seen in many medical image segmentation challenges requiring the segmentation of multiple classes simultaneously. This becomes particularly challenging for small or closely located organs. Secondly, segmentation quality heavily relies on the precision of the prompts, yet creating accurate prompts demands domain-specific expertise, often unavailable to non-expert users. These limitations diminish the practicality of prompt-based methods.

To overcome these limitations,  we propose to use an Auto Prompt Generator instead of positional encoding to represent the prompt. The whole process is illustrated in \Cref{fig2}C. Instead of using manually generated points or bounding boxes, we directly take the output feature map after the last block of attention and depth adapter operation. This APG follows a fully convolutional neural (FCN) based encoder-decoder design that resembles 3D UNet~\cite{ronneberger2015u}. This generator boasts a lightweight structure, leveraging 3D-based convolution operations, and can be effortlessly learned from scratch. This enables precise prompt generation tailored to different medical segmentation tasks. Notably, it eliminates the need for additional manually generated prompts, simplifying and expediting the multi-class medical image segmentation tasks.   

\subsection{Mask Decoder}

The SAM mask decoder was originally designed to be lightweight, using stacks of 2D convolution layers. In our updated version, we replaced these with 3D convolutions for direct 3D mask generation. While the original design works well for natural images with large, distinct objects, volumetric medical image segmentation benefits from U-shaped networks with multi-level skip connections~\cite{isensee2021nnu,tang2022self}. This is crucial for medical images, where objects are smaller and have less distinct boundaries, requiring higher-resolution details for better discrimination.

To address this need while keeping the design lightweight, we incorporate a multi-layer aggregation mechanism (MLAM)~\cite{zheng2021rethinking, gong20233dsam} in our decoder. We utilize the intermediate feature maps from stages 1-4 (as depicted in \Cref{fig2}A) of the image encoder, along with the prompt embedding from the APG, to enrich the mask feature map without compromising efficiency. For improved resolution detail, we upsample the mask feature map to match the original resolution, then concatenate it with the original image. This concatenated map is fused using another 3D convolution to generate the final mask. This method effectively combines high-resolution details with the original image data in the mask generation process. We have streamlined the original SAM by focusing solely on a targeted downstream task and omitting features like multi-task generation and ambiguity awareness. The backbone of the mask decoder primarily comprises lightweight 3D convolutional layers, which are known for their ease of optimization. Consequently, we train all decoder parameters from scratch.

\section{Experiments}
\label{sec:experiments}

\subsection{Datasets}
A total of 4 CT datasets, including 3 public and 1 institutional dataset, are used to evaluate the performance of models for the 3D multi-organ segmentation task.

\textbf{BTCV:} Beyond the Cranial Vault (BTCV) abdomen challenge dataset~\cite{landman2015miccai} includes 30 subjects with abdominal CT scans. In this dataset, 13 organs are annotated by interpreters under the supervision of Vanderbilt University Medical Center radiologists. The multi-organ segmentation task is framed as a 13-class segmentation, which includes large organs such as liver, spleen, kidneys, and stomach; vascular tissues of esophagus, aorta, inferior vena cava, splenic, and portal veins; small anatomies of gallbladder, pancreas, and adrenal glands. For the subjects, 24 scans are for training and 6 for testing. 

\textbf{AMOS:} We mainly focus on the CT parts of the AMOS dataset~\cite{ji2022amos}. The publicly accessible AMOS-CT dataset~\cite{ji2022amos} consists of 200 multi-contrast abdominal CT scans for training and 100 for testing. These scans are annotated for 15 anatomies,  including spleen, right kidney, left kidney, gallbladder, esophagus, liver, stomach, aorta, inferior vena cava, pancreas, right adrenal gland, left adrenal gland, duodenum, bladder, and prostate/uterus.

\textbf{CT-ORG:} CT Organ Segmentation Dataset (CT-ORG) \cite{rister2020ct} comprises 100 CT scans, each of which includes ground-truth contours of 5 different organs (lung, bones, liver, kidney and bladder). Of these, the first 19 CT scans are exclusively reserved for testing purposes whereas the remaining 81 scans are used for training.

\textbf{Institutional Pelvic:} In addition to public datasets, we also test the performance of the segmentation models on a private dataset. An institutional pelvic dataset of 300 prostate cancer patients is retrospectively selected, and each contains 3 manually generated contours (prostate, bladder, and rectum). The 300 cases are randomly split into a training set of 225 cases, a validation set of 30, and a testing set of 45.

\subsection{Implementation Details}
The AutoProSAM model is trained using the AdamW optimizer with a warm-up exponential decay scheduler of 200 epochs, in which we incorporate the first 5 epochs for warmup. The segmentation experiments use a batch size of 1 per GPU with a patch size of $128 \times 128 \times 128$. Default initial learning rate of $5 e^{-4}$, momentum of 0.9 and decay of $1 e^{-5}$. The framework is implemented in MONAI 1.2 and PyTorch 2.0. Models are trained on a server with eight NVIDIA A100 cards. All the baseline models are trained from scratch following their default settings. For each dataset, we perform interpolation for a fixed voxel spacing over all scans within this dataset, then intensity normalization, sampling over positive and negative patches, and finally on-the-fly data augmentation with flip, rotation, intensity scaling, and intensity shifting. Foreground and background patches are randomly sampled at a $1:1$ ratio. We select the best model with the highest Dice metric on the validation dataset or the final epoch's model for a dataset without a predefined validation split, with empirical evidence of negligible gain via a 5-fold cross-validation strategy. Baseline models are trained from scratch following their default settings. For inference, an overlapping area ratio of 0.75 is applied via the sliding window strategy. 

We provide detailed pre-processing information about the interpolation to a fixed voxel spacing, HU ranges, and intensity normalization in \Cref{tab: preprocessing}.

\begin{table}[h!]
\scriptsize
\begin{tabular}{|c|c|c|c|}
\hline
\textbf{Dataset} & \textbf{Voxel Spacing (mm)} & \textbf{HU Range} & \textbf{Normalization} \\ \hline
BTCV & $1.0\times 1.0\times1.5$ & $[-125, 275]$ & $[0, 1]$  \\ \hline
\multirow{2}{*}{AMOS} & \multirow{2}{*}{$1.0\times 1.0\times1.5$} & $\multirow{2}{*}{[-991, 362]}$ & Subtract 50, \\ 
 & & & divide by 141 \\ 
\hline
CT-ORG & Isotropic 2.0 & $[-1000, 1000]$ & $[-1, 1]$ \\ \hline
Institutional & \multirow{2}{*}{Isotropic 1.5} & \multirow{2}{*}{$[-50, 150]$} & \multirow{2}{*}{$[0, 1]$}  \\ 
Pelvic & & &  \\ \hline

\end{tabular}
\caption{Pre-processing settings for each dataset.}
\label{tab: preprocessing}
\end{table}

\noindent
A sampling of $128 \times 128 \times 128$ voxel patches is done for all four datasets, with a positive-to-negative patches ratio of 1:1 sampling for AMOS, CT-ORG, and BTCV datasets.

\subsection{Loss Function}
For training the AutoProSAM (as shown in \Cref{fig2}A), a combination of Dice loss and Cross-Entropy loss is used to assess the alignment between the predicted mask and the ground truth on a pixel-wise basis. The objective function for the segmentation head is defined as follows:
\begin{equation} \label{eqn_1}
    \mathcal{L}_\mathrm{Seg} = \mathcal{L}_\mathrm{Dice} (\hat{p}_i, g_i) + \mathcal{L}_\mathrm{CE}(\hat{p}_i, g_i),
\end{equation}
where $\hat{p}_i$ represents the predicted voxel probabilities from the main task, and $g_i$ represents the ground truth mask for an input volume $i$. The predicted probabilities, $\hat{p}_i$, result from applying the AutoProSAM to the input 3D volume for the main task.

\subsection{Evaluation Metrics}
We utilize the Dice similarity coefficient (Dice) and Normalized Surface Distance (NSD)~\cite{nikolov2018deep} as metrics to evaluate the segmentation performance. Dice metric measures the extent of agreement between two volumes with the following formula:
\begin{equation} \label{eq_dice}
\mathrm{Dice} = \frac{2\sum_{i=1}^I g_i \hat{g}_i}{\sum_{i=1}^I g_i + \sum_{i=1}^I \hat{g}_i},
\end{equation}
where $g$ and $\hat{g}$ denote the ground truth and the predicted voxel values. The NSD metric quantifies the overlap between ground truth and predicted surfaces (with a fixed tolerance). For both metrics, 0 denotes the worst match between two structures, and 100\% means the perfect match.

\label{sec: r_n_d}
\begin{figure*}[ht!]
\centering
\includegraphics[width=0.95\textwidth]{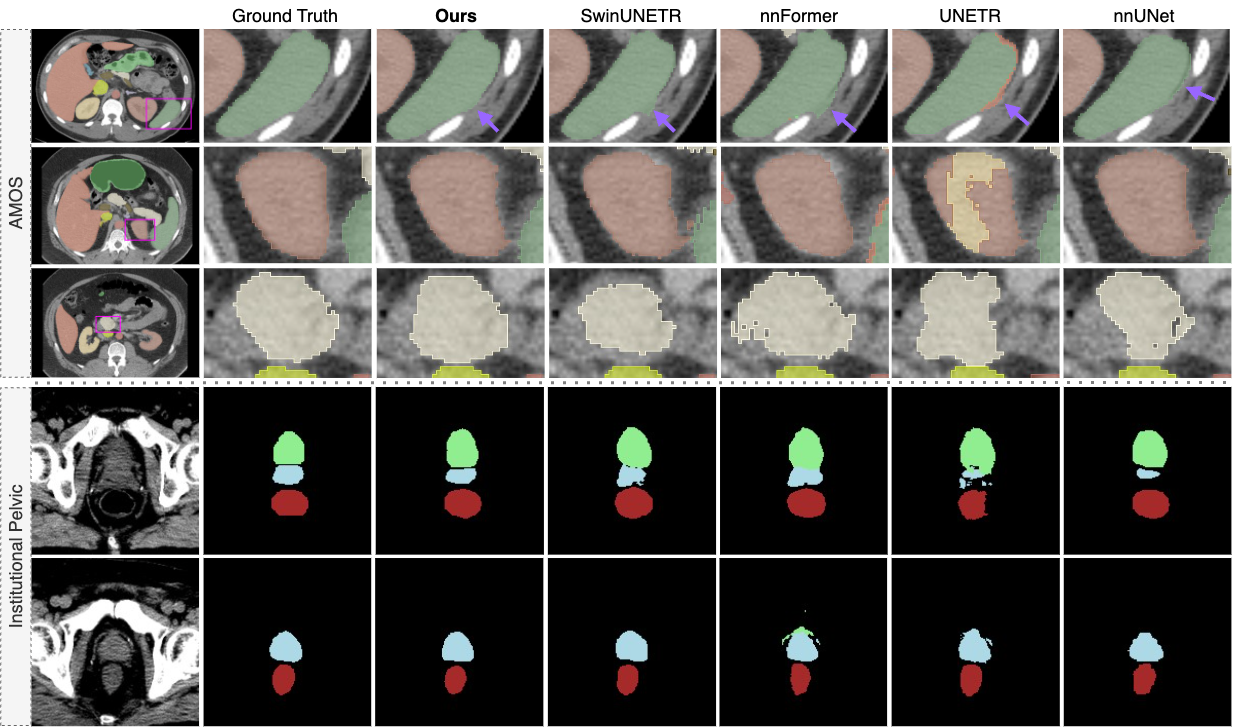} 
\caption{Qualitative visualizations compare our \textbf{AutoProSAM} with baseline methods using three subjects from public datasets (Rows 1-3) and two subjects from the private Institutional Pelvic dataset (Rows 4-5). Enhanced areas in these visualizations illustrate improvements in segmenting the left kidney (light red) and pancreas (beige). Additionally, segmentation masks are shown for the prostate (blue), bladder (green), and rectum (red).}
\label{fig3}
\end{figure*} 
 
\section{Results \& Discussion}
\label{sec:r_n_d}

\begin{table*}[ht!]
        \setlength{\tabcolsep}{3pt}
        \centering
        \small
	\begin{tabular}{|l|cc|cc|cc|cc|} \hline
            \multirow{2}{*}{Model}  &\multicolumn{2}{c|}{BTCV} &\multicolumn{2}{c|}{AMOS}  &\multicolumn{2}{c|}{CT-ORG} &\multicolumn{2}{c|}{Institutional Pelvic} \\
		&mDice  $\uparrow$ & mNSD  $\uparrow$ &mDice $\uparrow$ &mNSD $\uparrow$ &mDice $\uparrow$ & mNSD $\uparrow$ &mDice $\uparrow$ &mNSD $\uparrow$  \\ \hline\hline
		nnUNet~\cite{isensee2021nnu}   & 84.34 & 73.21 & 87.43 & 77.12  & \textbf{85.51} & 75.60& 88.04 & 83.87\\
        AttUNet~\cite{wang2021attu}   & 83.67 & 74.35 & 83.75 & 72.23  & 85.13 & 74.64& 88.06 & 84.12\\
        UNet++~\cite{zhou2019unet++}   & 79.33 & 72.84 & 84.32 & 73.34  & 84.87 &74.21& 87.53 & 82.65\\
		nnFormer~\cite{zhou2023nnformer}  &83.51 &71.65 & 84.52 & 70.06  & 82.75 & 70.31 & 87.60 & 82.13\\ 
		UNETR~\cite{hatamizadeh2022unetr}  &85.47 &74.35 & 77.24 & 60.58  & 83.13 & 71.56 & 86.68 & 81.54\\
		SwinUNETR~\cite{tang2022self} &86.58 &75.26 & 86.19 & 74.83 & 84.34 & 72.32 & 89.12 & 83.54\\ 
            \textbf{Ours}  &\textbf{87.15} &\textbf{78.83*} &\textbf{88.65*} & \textbf{79.41*} & 85.12 &\textbf{76.37*} &\textbf{91.30*} & \textbf{84.35*}\\ \hline
	\end{tabular}
	\caption{Comparison of the overall performance between four SOTAs and our \textbf{AutoProSAM} on four datasets. The best results are highlighted in bold font. (*: $p < 0.01$, with Wilcoxon signed-rank test to all SOTAs)}
    \vspace{-5mm}
	\label{tab_overall}
\end{table*}

In this section, we would like to evaluate the performance of our proposed \textbf{AutoProSAM} concerning both the task-specific SOTAs and the SAM-adopting-based methods.  The effectiveness of different modules for our designs will be evaluated as well.

\subsection{Comparison with SOTAs}

We extensively compare our model with the SOTA 3D medical image segmentation approaches, including the most recent Transformer-based methods including UNETR~\cite{hatamizadeh2022unetr}, SwinUNETR~\cite{tang2022self}, and nnFormer~\cite{zhou2023nnformer}, as well as CNN-based methods such as nnUNet~\cite{isensee2021nnu}, AttUNet~\cite{wang2021attu}, and UNet++~\cite{zhou2019unet++}. As indicated in \Cref{tab_overall}, we observe that the proposed \textbf{AutoProSAM} generally outperforms other SOTA methods in both Dice and NSD metrics across all four datasets. 

Significant improvements are particularly noticeable in the BTCV dataset, with up to a 3\% increase in the average Dice score and a 3\% to 7\% improvement in the average NSD metric. As the training sample size increases for the AMOS dataset compared to BTCV, the AutoProSAM performs even better, achieving up to a 14\% improvement in Dice and a 2\% to 19\% improvement in NSD. For the CT-ORG dataset, our model achieves the highest NSD and a competitive Dice score compared to other baselines. Notably, the AutoProSAM consistently demonstrates a robust performance in real-world scenarios, as seen in its consistently superior performance on the private Institutional Pelvic dataset.

The superiority of our AutoProSAM model is further highlighted by qualitative results. As depicted in \Cref{fig3}, qualitative comparisons of the predicted masks from various segmentation models reveal that our AutoProSAM yields visually superior mask predictions, particularly in terms of more accurate boundary delineation, when compared to its SOTA counterparts. In summary, while the original SAM displayed relatively weaker performance in medical image segmentation tasks as compared to SOTA methods~\cite{mazurowski2023segment}, our adapted design registers significant enhancements and generalization to 3D medical image segmentation. 

\begin{figure}[h]
\centering
\includegraphics[width=0.47\textwidth]{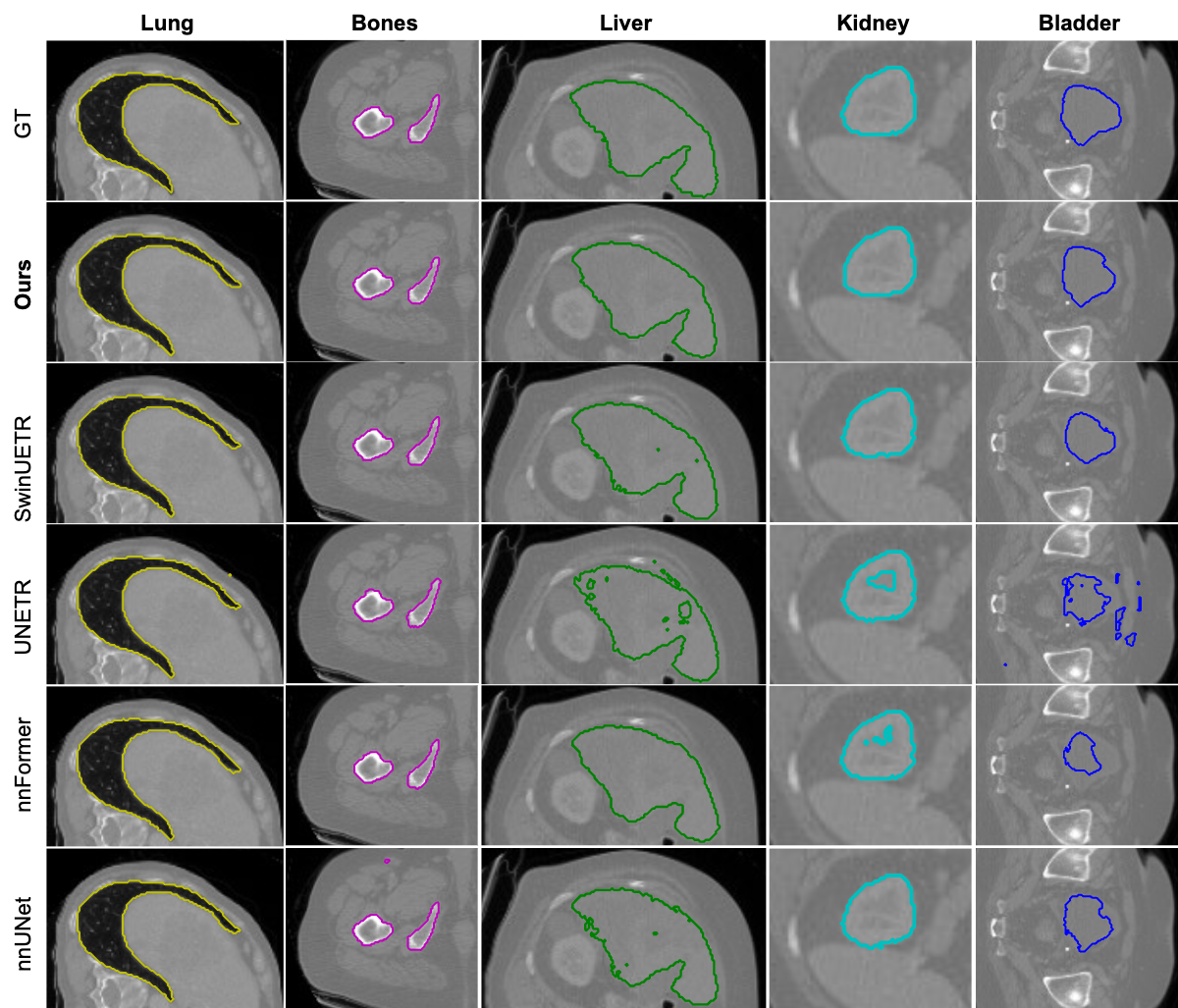}
\caption{Qualitative visualizations compare our AutoProSAM with baseline methods over the CT-ORG dataset.}
\label{fig_s1}
\vspace{-5mm}
\end{figure}

We also randomly selected samples from the CT-ORG dataset for more visual comparisons of different methods. As depicted in \Cref{fig_s1}, all methods generally achieve precise prediction masks for organs that are relatively easy to segment, such as lungs and bones. However, our AutoProSAM also provides more consistently accurate predictions than other baselines for organs like the liver, kidney, and bladder.

\subsection{Comparison with SAM-based Methods}
We further compare our method with existing SAM-based methods for multi-class medical image segmentation, including the original SAM and MedSAM~\cite{ma2023segment}. MedSAM can be implemented with either full fine-tuning or partial fine-tuning, where only the prompt encoder and mask decoder are updated using point prompts. Other adaptation methods were not considered, as they either require collecting a large number of medical images for fine-tuning (\eg, Medical-SAM-Adapter~\cite{wu2023medical}, SAM-Med2D~\cite{cheng2023sam}, SAM-Med3D~\cite{wang2023sam}), or their models are too huge, where comparing would not be fair (Medical SAM Adapter~\cite{wu2023medical} has tunable parameters approximately 336 million compared to our model of ten times fewer), or some models do not focus on organ segmentation (\eg, 3DSAM-Adapter~\cite{gong20233dsam} for tumor segmentation, MA-SAM~\cite{chen2024ma} focuses on modality-agnostic performance for medical images), or some models are 2D models working with 2D slices (\eg, H-SAM~\cite{cheng2024unleashing} and SimTxtSeg~\cite{xie2024simtxtseg}). Furthermore, we observe that many SAM-based works rely on ground truth images during the inference stage to generate prompts, which provides an unfair performance advantage. In contrast, our method, APG, does not depend on ground truth images, making it significantly more robust. 

\begin{table}[t!]
        \setlength{\tabcolsep}{3pt}
        \centering
	\small
	\scalebox{0.95}{
        \begin{tabular}{|l|c|cc|cc|} \hline
		\multirow{2}{*}{Model} &{Tuning}  &\multicolumn{2}{c|}{BTCV} &\multicolumn{2}{c|}{AMOS}  \\
		& Option& mDice  $\uparrow$&mNSD  $\uparrow$ & mDice $\uparrow$ & mNSD $\uparrow$\\ \hline\hline
		  SAM~\cite{kirillov2023segment}   & None  & 54.86 & - & 49.31 & -  \\
            MedSAM~\cite{ma2023segment} & P\&M  & 80.65 & 66.82 & 70.28 & 59.61  \\ 
            MedSAM~\cite{ma2023segment} & Full & 84.57 & 73.76 & 83.10 & 70.26  \\
            \textbf{Ours} & P\&M  &\textbf{87.15} &\textbf{78.83} &\textbf{88.65} & \textbf{79.41}\\ \hline
	\end{tabular}
        }
	\caption{Comparison with SAM-based methods, where the best results are highlighted in bold. `P\&M' indicates only fine-tuning the prompt encoder and mask decoder, `Full' denotes full fine-tuning, and `None' represents no fine-tuning. }
        \vspace{-5mm}
	\label{tab_adaption}
\end{table}

The outcomes are meticulously outlined in \Cref{tab_adaption}, underscoring how our adaptation strategy surpasses all existing methods. Notably, our approach outshines the second-best technique by a margin of 3\% in terms of BTCV segmentation Dice, 5\% for NSD, and 6\% concerning AMOS segmentation Dice, accompanied by a substantial 9\% for NSD. Impressively, it even outperforms the complete fine-tuning variant of MedSAM, even when considering parameter-efficient fine-tuning. These results effectively validate our hypothesis that parameters pre-trained on 2D images can be effectively harnessed to grasp 3D spatial features with only minor adjustments. Moreover, our approach of treating all dimensions equivalently emerges as a superior strategy compared to interpreting the depth dimension as a distinct group in the context of medical image segmentation. 

\subsection{Ablation Study}

 \noindent \textbf{{Effects of APG and MLAM}}
The APG in AutoProSAM refines feature maps from the final stage of the encoder, similar to how SAM's prompt encoder processes manual prompts. Without APG, these maps go directly to the mask decoder. Additionally, the Multi-Layer Aggregation Mechanism (MLAM) integrates information from various stages into the final mask decoder. A comparative analysis on the BTCV dataset, shown in \Cref{tab_ab1}, demonstrates significant improvements in Dice and NSD metrics when APG and MLAM are used, highlighting their critical role in improving segmentation performance.

\begin{table}[t!]
        \setlength{\tabcolsep}{3pt}
        \centering
	\scriptsize
	\begin{tabular}{|cc|cc|} \hline
		APG & MLAM &mDice  $\uparrow$ & mNSD  $\uparrow$ \\ \hline\hline
            \cmark & \cmark  & \textbf{87.15} & \textbf{78.83} \\ 
            \xmark & \cmark& 85.23 & 74.12\\ 
            \cmark & \xmark & 84.37 & 73.54\\ 
            \xmark & \xmark & 81.14 & 70.33\\ \hline
	\end{tabular}
 
        \caption{Ablation study evaluating the effectiveness of the Automated Prompt Generator (APG) and Multi-Level Attention Mechanism (MLAM) designs in our \textbf{AutoProSAM}.}
        \vspace{-5mm}
	\label{tab_ab1}
\end{table}

 \noindent \textbf{{Computational Requirements Analysis}}
As demonstrated in \Cref{tab_s1}, our AutoProSAM, despite incorporating a heavy ViT encoder, achieves an inference speed only marginally slower than that of nnUNet \cite{isensee2021nnu} and SwinUNETR \cite{tang2022self}. Notably, it significantly outperforms both UNETR \cite{hatamizadeh2022unetr} and nnFormer \cite{zhou2023nnformer} in terms of speed, highlighting its remarkable efficiency among these SOTA methods. Furthermore, AutoProSAM has the fewest tunable parameters during training, making it the most computationally feasible among the benchmark models.

\begin{table}[h]
\setlength{\tabcolsep}{3pt}
\centering
\scriptsize
\begin{tabular}{|l|c|c|}
\hline
\multirow{2}{*}{Model} & Tuned &Inference Time \\
&Params. (M) &(sec./scan)\\ \hline\hline
nnUNet~\cite{isensee2021nnu}  &31.18& 3.74\\
nnFormer~\cite{zhou2023nnformer}  & 150.14& 11.73\\
UNETR~\cite{hatamizadeh2022unetr}  & 93.02& 58.93\\
SwinUNETR~\cite{tang2022self}  & 62.83& 4.24\\
AutoProSAM (ours)  &26.53*& 7.06\\
\hline
\end{tabular}
\caption{Averaged inference time for different models in BTCV experiments with a patch size of $128 \times 128 \times 128$.*Note: this may change slightly according to your exact setting of implementation.}
\vspace{-3mm}
\label{tab_s1}
\end{table}
 
 \noindent \textbf{{Effectiveness for MRI Modality and More Organs}}
 To evaluate AutoProSAM, we compared its average performance on AMOS MRI~\cite{ji2022amos} and MM-WHS CT (heart substructures)~\cite{zhuang2019evaluation} datasets against benchmark models. The average dice scores, shown in \Cref{tab:amos_mmwhs_comparison}, demonstrate AutoProSAM’s superiority across various configurations.

\begin{table}[htbp]
    \scriptsize
    \centering
    \begin{tabular}{|l|c|c|}
        \hline
        Model & AMOS MRI & MM-WHS CT \\ \hline\hline
        nnFormer & 80.6 & 79.6 \\ 
        UNETR & 75.3 & 85.8 \\ 
        SwinUNETR & 75.7 & 86.1 \\ 
        \textbf{AutoProSAM} & \textbf{80.8} & \textbf{90.9} \\ \hline
    \end{tabular}
    \caption{Comparisons on AMOS MRI~\cite{ji2022amos} and MM-WHS CT~\cite{zhuang2019evaluation} datasets on average Dice.}
    
    \vspace{-0.20in}
    \label{tab:amos_mmwhs_comparison}
\end{table}

\section{Conclusion}
\label{sec:conclusion}

In this paper, we introduced AutoProSAM, a novel approach to enhance SAM for 3D multi-organ medical image segmentation. By adapting SAM from 2D natural images to 3D medical images, our method addresses domain gaps and spatial disparities through parameter-efficient fine-tuning and an Auto Prompt Generator (APG) that automates prompt creation, removing manual input. Extensive experiments on public and private CT-based datasets demonstrated AutoProSAM's superior performance over state-of-the-art models, with higher Dice and NSD scores. Ablation studies also confirmed the critical role of APG and the Multi-Layer Aggregation Mechanism (MLAM) in boosting segmentation accuracy. 

{\small
\bibliographystyle{ieee_fullname}
\bibliography{egbib}

\begin{thebibliography}{10}\itemsep=-1pt

\bibitem{chen2024ma}
Cheng Chen, Juzheng Miao, Dufan Wu, Aoxiao Zhong, Zhiling Yan, Sekeun Kim, Jiang Hu, Zhengliang Liu, Lichao Sun, Xiang Li, et~al.
\newblock Ma-sam: Modality-agnostic sam adaptation for 3d medical image segmentation.
\newblock {\em Medical Image Analysis}, page 103310, 2024.

\bibitem{cheng2023sam}
Junlong Cheng, Jin Ye, Zhongying Deng, Jianpin Chen, Tianbin Li, Haoyu Wang, Yanzhou Su, Ziyan Huang, Jilong Chen, Lei Jiang, et~al.
\newblock Sam-med2d.
\newblock {\em arXiv preprint arXiv:2308.16184}, 2023.

\bibitem{cheng2024unleashing}
Zhiheng Cheng, Qingyue Wei, Hongru Zhu, Yan Wang, Liangqiong Qu, Wei Shao, and Yuyin Zhou.
\newblock Unleashing the potential of sam for medical adaptation via hierarchical decoding.
\newblock In {\em Proceedings of the IEEE/CVF Conference on Computer Vision and Pattern Recognition}, pages 3511--3522, 2024.

\bibitem{colbert2024repurposing}
Zachery~Morton Colbert, Daniel Arrington, Matthew Foote, Jonas G{\aa}rding, Dominik Fay, Michael Huo, Mark Pinkham, and Prabhakar Ramachandran.
\newblock Repurposing traditional u-net predictions for sparse sam prompting in medical image segmentation.
\newblock {\em Biomedical Physics \& Engineering Express}, 10(2):025004, 2024.

\bibitem{ding2023parameter}
Ning Ding, Yujia Qin, Guang Yang, Fuchao Wei, Zonghan Yang, Yusheng Su, Shengding Hu, Yulin Chen, Chi-Min Chan, Weize Chen, et~al.
\newblock Parameter-efficient fine-tuning of large-scale pre-trained language models.
\newblock {\em Nature Machine Intelligence}, 5(3):220--235, 2023.

\bibitem{dosovitskiy2020image}
Alexey Dosovitskiy, Lucas Beyer, Alexander Kolesnikov, Dirk Weissenborn, Xiaohua Zhai, Thomas Unterthiner, Mostafa Dehghani, Matthias Minderer, Georg Heigold, Sylvain Gelly, et~al.
\newblock An image is worth 16x16 words: Transformers for image recognition at scale.
\newblock {\em arXiv preprint arXiv:2010.11929}, 2020.

\bibitem{gao2023desam}
Yifan Gao, Wei Xia, Dingdu Hu, and Xin Gao.
\newblock Desam: Decoupling segment anything model for generalizable medical image segmentation.
\newblock {\em arXiv preprint arXiv:2306.00499}, 2023.

\bibitem{goldblum2023battle}
Micah Goldblum, Hossein Souri, Renkun Ni, Manli Shu, Viraj Prabhu, Gowthami Somepalli, Prithvijit Chattopadhyay, Mark Ibrahim, Adrien Bardes, Judy Hoffman, et~al.
\newblock Battle of the backbones: A large-scale comparison of pretrained models across computer vision tasks.
\newblock {\em arXiv preprint arXiv:2310.19909}, 2023.

\bibitem{gong20233dsam}
Shizhan Gong, Yuan Zhong, Wenao Ma, Jinpeng Li, Zhao Wang, Jingyang Zhang, Pheng-Ann Heng, and Qi Dou.
\newblock 3dsam-adapter: Holistic adaptation of sam from 2d to 3d for promptable medical image segmentation.
\newblock {\em arXiv preprint arXiv:2306.13465}, 2023.

\bibitem{guo2020parameter}
Demi Guo, Alexander~M Rush, and Yoon Kim.
\newblock Parameter-efficient transfer learning with diff pruning.
\newblock {\em arXiv preprint arXiv:2012.07463}, 2020.

\bibitem{hatamizadeh2022unetr}
Ali Hatamizadeh, Yucheng Tang, Vishwesh Nath, Dong Yang, Andriy Myronenko, Bennett Landman, Holger~R Roth, and Daguang Xu.
\newblock Unetr: Transformers for 3d medical image segmentation.
\newblock In {\em Proceedings of the IEEE/CVF winter conference on applications of computer vision}, pages 574--584, 2022.

\bibitem{hu2021lora}
Edward~J Hu, Yelong Shen, Phillip Wallis, Zeyuan Allen-Zhu, Yuanzhi Li, Shean Wang, Lu Wang, and Weizhu Chen.
\newblock Lora: Low-rank adaptation of large language models.
\newblock {\em arXiv preprint arXiv:2106.09685}, 2021.

\bibitem{isensee2021nnu}
Fabian Isensee, Paul~F Jaeger, Simon~AA Kohl, Jens Petersen, and Klaus~H Maier-Hein.
\newblock nnu-net: a self-configuring method for deep learning-based biomedical image segmentation.
\newblock {\em Nature methods}, 18(2):203--211, 2021.

\bibitem{ji2022amos}
Yuanfeng Ji, Haotian Bai, Chongjian Ge, Jie Yang, Ye Zhu, Ruimao Zhang, Zhen Li, Lingyan Zhanng, Wanling Ma, Xiang Wan, et~al.
\newblock Amos: A large-scale abdominal multi-organ benchmark for versatile medical image segmentation.
\newblock {\em Advances in Neural Information Processing Systems}, 35:36722--36732, 2022.

\bibitem{jia2022visual}
Menglin Jia, Luming Tang, Bor-Chun Chen, Claire Cardie, Serge Belongie, Bharath Hariharan, and Ser-Nam Lim.
\newblock Visual prompt tuning.
\newblock In {\em European Conference on Computer Vision}, pages 709--727. Springer, 2022.

\bibitem{jing2020self}
Longlong Jing and Yingli Tian.
\newblock Self-supervised visual feature learning with deep neural networks: A survey.
\newblock {\em IEEE transactions on pattern analysis and machine intelligence}, 43(11):4037--4058, 2020.

\bibitem{kirillov2023segment}
Alexander Kirillov, Eric Mintun, Nikhila Ravi, Hanzi Mao, Chloe Rolland, Laura Gustafson, Tete Xiao, Spencer Whitehead, Alexander~C Berg, Wan-Yen Lo, et~al.
\newblock Segment anything.
\newblock {\em arXiv preprint arXiv:2304.02643}, 2023.

\bibitem{landman2015miccai}
Bennett Landman, Zhoubing Xu, J Igelsias, Martin Styner, T Langerak, and Arno Klein.
\newblock Miccai multi-atlas labeling beyond the cranial vault--workshop and challenge.
\newblock In {\em Proc. MICCAI Multi-Atlas Labeling Beyond Cranial Vault—Workshop Challenge}, volume~5, page~12, 2015.

\bibitem{lei2023medlsam}
Wenhui Lei, Xu Wei, Xiaofan Zhang, Kang Li, and Shaoting Zhang.
\newblock Medlsam: Localize and segment anything model for 3d medical images.
\newblock {\em arXiv preprint arXiv:2306.14752}, 2023.

\bibitem{li2023focalunetr}
Chengyin Li, Yao Qiang, Rafi~Ibn Sultan, Hassan Bagher-Ebadian, Prashant Khanduri, Indrin~J Chetty, and Dongxiao Zhu.
\newblock Focalunetr: A focal transformer for boundary-aware prostate segmentation using ct images.
\newblock In {\em International Conference on Medical Image Computing and Computer-Assisted Intervention}, pages 592--602. Springer, 2023.

\bibitem{lin2014microsoft}
Tsung-Yi Lin, Michael Maire, Serge Belongie, James Hays, Pietro Perona, Deva Ramanan, Piotr Doll{\'a}r, and C~Lawrence Zitnick.
\newblock Microsoft coco: Common objects in context.
\newblock In {\em Computer Vision--ECCV 2014: 13th European Conference, Zurich, Switzerland, September 6-12, 2014, Proceedings, Part V 13}, pages 740--755. Springer, 2014.

\bibitem{liu2023pre}
Pengfei Liu, Weizhe Yuan, Jinlan Fu, Zhengbao Jiang, Hiroaki Hayashi, and Graham Neubig.
\newblock Pre-train, prompt, and predict: A systematic survey of prompting methods in natural language processing.
\newblock {\em ACM Computing Surveys}, 55(9):1--35, 2023.

\bibitem{ma2023segment}
Jun Ma and Bo Wang.
\newblock Segment anything in medical images.
\newblock {\em arXiv preprint arXiv:2304.12306}, 2023.

\bibitem{mazurowski2023segment}
Maciej~A Mazurowski, Haoyu Dong, Hanxue Gu, Jichen Yang, Nicholas Konz, and Yixin Zhang.
\newblock Segment anything model for medical image analysis: an experimental study.
\newblock {\em Medical Image Analysis}, page 102918, 2023.

\bibitem{min2021recent}
Bonan Min, Hayley Ross, Elior Sulem, Amir Pouran~Ben Veyseh, Thien~Huu Nguyen, Oscar Sainz, Eneko Agirre, Ilana Heintz, and Dan Roth.
\newblock Recent advances in natural language processing via large pre-trained language models: A survey.
\newblock {\em ACM Computing Surveys}, 2021.

\bibitem{na2024segment}
Saiyang Na, Yuzhi Guo, Feng Jiang, Hehuan Ma, and Junzhou Huang.
\newblock Segment any cell: A sam-based auto-prompting fine-tuning framework for nuclei segmentation.
\newblock {\em arXiv preprint arXiv:2401.13220}, 2024.

\bibitem{nikolov2018deep}
Stanislav Nikolov, Sam Blackwell, Alexei Zverovitch, Ruheena Mendes, Michelle Livne, Jeffrey De~Fauw, Yojan Patel, Clemens Meyer, Harry Askham, Bernardino Romera-Paredes, et~al.
\newblock Deep learning to achieve clinically applicable segmentation of head and neck anatomy for radiotherapy.
\newblock {\em arXiv preprint arXiv:1809.04430}, 2018.

\bibitem{oquab2023dinov2}
Maxime Oquab, Timoth{\'e}e Darcet, Th{\'e}o Moutakanni, Huy Vo, Marc Szafraniec, Vasil Khalidov, Pierre Fernandez, Daniel Haziza, Francisco Massa, Alaaeldin El-Nouby, et~al.
\newblock Dinov2: Learning robust visual features without supervision.
\newblock {\em arXiv preprint arXiv:2304.07193}, 2023.

\bibitem{pan2022st}
Junting Pan, Ziyi Lin, Xiatian Zhu, Jing Shao, and Hongsheng Li.
\newblock St-adapter: Parameter-efficient image-to-video transfer learning.
\newblock {\em Advances in Neural Information Processing Systems}, 35:26462--26477, 2022.

\bibitem{pandey2023comprehensive}
Sumit Pandey, Kuan-Fu Chen, and Erik~B Dam.
\newblock Comprehensive multimodal segmentation in medical imaging: Combining yolov8 with sam and hq-sam models.
\newblock In {\em Proceedings of the IEEE/CVF International Conference on Computer Vision}, pages 2592--2598, 2023.

\bibitem{qiang2023interpretability}
Yao Qiang, Chengyin Li, Prashant Khanduri, and Dongxiao Zhu.
\newblock Interpretability-aware vision transformer.
\newblock {\em arXiv preprint arXiv:2309.08035}, 2023.

\bibitem{radford2021learning}
Alec Radford, Jong~Wook Kim, Chris Hallacy, Aditya Ramesh, Gabriel Goh, Sandhini Agarwal, Girish Sastry, Amanda Askell, Pamela Mishkin, Jack Clark, et~al.
\newblock Learning transferable visual models from natural language supervision.
\newblock In {\em International conference on machine learning}, pages 8748--8763. PMLR, 2021.

\bibitem{rister2020ct}
Blaine Rister, Darvin Yi, Kaushik Shivakumar, Tomomi Nobashi, and Daniel~L Rubin.
\newblock Ct-org, a new dataset for multiple organ segmentation in computed tomography.
\newblock {\em Scientific Data}, 7(1):381, 2020.

\bibitem{ronneberger2015u}
Olaf Ronneberger, Philipp Fischer, and Thomas Brox.
\newblock U-net: Convolutional networks for biomedical image segmentation.
\newblock In {\em Medical Image Computing and Computer-Assisted Intervention--MICCAI 2015: 18th International Conference, Munich, Germany, October 5-9, 2015, Proceedings, Part III 18}, pages 234--241. Springer, 2015.

\bibitem{shaharabany2023autosam}
Tal Shaharabany, Aviad Dahan, Raja Giryes, and Lior Wolf.
\newblock Autosam: Adapting sam to medical images by overloading the prompt encoder.
\newblock {\em arXiv preprint arXiv:2306.06370}, 2023.

\bibitem{tang2022self}
Yucheng Tang, Dong Yang, Wenqi Li, Holger~R Roth, Bennett Landman, Daguang Xu, Vishwesh Nath, and Ali Hatamizadeh.
\newblock Self-supervised pre-training of swin transformers for 3d medical image analysis.
\newblock In {\em Proceedings of the IEEE/CVF Conference on Computer Vision and Pattern Recognition}, pages 20730--20740, 2022.

\bibitem{wang2023sam}
Haoyu Wang, Sizheng Guo, Jin Ye, Zhongying Deng, Junlong Cheng, Tianbin Li, Jianpin Chen, Yanzhou Su, Ziyan Huang, Yiqing Shen, et~al.
\newblock Sam-med3d.
\newblock {\em arXiv preprint arXiv:2310.15161}, 2023.

\bibitem{wang2021attu}
Sihan Wang, Lei Li, and Xiahai Zhuang.
\newblock Attu-net: attention u-net for brain tumor segmentation.
\newblock In {\em International MICCAI brainlesion workshop}, pages 302--311. Springer, 2021.

\bibitem{wang2023med}
Wenxuan Wang, Jiachen Shen, Chen Chen, Jianbo Jiao, Yan Zhang, Shanshan Song, and Jiangyun Li.
\newblock Med-tuning: Exploring parameter-efficient transfer learning for medical volumetric segmentation.
\newblock {\em arXiv preprint arXiv:2304.10880}, 2023.

\bibitem{wang2022contrastive}
Xiao Wang and Guo-Jun Qi.
\newblock Contrastive learning with stronger augmentations.
\newblock {\em IEEE transactions on pattern analysis and machine intelligence}, 45(5):5549--5560, 2022.

\bibitem{wu2023medical}
Junde Wu, Rao Fu, Huihui Fang, Yuanpei Liu, Zhaowei Wang, Yanwu Xu, Yueming Jin, and Tal Arbel.
\newblock Medical sam adapter: Adapting segment anything model for medical image segmentation.
\newblock {\em arXiv preprint arXiv:2304.12620}, 2023.

\bibitem{xie2024simtxtseg}
Yuxin Xie, Tao Zhou, Yi Zhou, and Geng Chen.
\newblock Simtxtseg: Weakly-supervised medical image segmentation with simple text cues.
\newblock {\em arXiv preprint arXiv:2406.19364}, 2024.

\bibitem{zaken2021bitfit}
Elad~Ben Zaken, Shauli Ravfogel, and Yoav Goldberg.
\newblock Bitfit: Simple parameter-efficient fine-tuning for transformer-based masked language-models.
\newblock {\em arXiv preprint arXiv:2106.10199}, 2021.

\bibitem{zhang2023comprehensive}
Chunhui Zhang, Li Liu, Yawen Cui, Guanjie Huang, Weilin Lin, Yiqian Yang, and Yuehong Hu.
\newblock A comprehensive survey on segment anything model for vision and beyond.
\newblock {\em arXiv preprint arXiv:2305.08196}, 2023.

\bibitem{zhang2023customized}
Kaidong Zhang and Dong Liu.
\newblock Customized segment anything model for medical image segmentation.
\newblock {\em arXiv preprint arXiv:2304.13785}, 2023.

\bibitem{SAM4MIS}
Yichi Zhang and Rushi Jiao.
\newblock How segment anything model (sam) boost medical image segmentation?
\newblock {\em arXiv preprint arXiv:2305.03678}, 2023.

\bibitem{zheng2021rethinking}
Sixiao Zheng, Jiachen Lu, Hengshuang Zhao, Xiatian Zhu, Zekun Luo, Yabiao Wang, Yanwei Fu, Jianfeng Feng, Tao Xiang, Philip~HS Torr, et~al.
\newblock Rethinking semantic segmentation from a sequence-to-sequence perspective with transformers.
\newblock In {\em Proceedings of the IEEE/CVF conference on computer vision and pattern recognition}, pages 6881--6890, 2021.

\bibitem{zhou2017scene}
Bolei Zhou, Hang Zhao, Xavier Puig, Sanja Fidler, Adela Barriuso, and Antonio Torralba.
\newblock Scene parsing through ade20k dataset.
\newblock In {\em Proceedings of the IEEE conference on computer vision and pattern recognition}, pages 633--641, 2017.

\bibitem{zhou2023nnformer}
Hong-Yu Zhou, Jiansen Guo, Yinghao Zhang, Xiaoguang Han, Lequan Yu, Liansheng Wang, and Yizhou Yu.
\newblock nnformer: Volumetric medical image segmentation via a 3d transformer.
\newblock {\em IEEE Transactions on Image Processing}, 2023.

\bibitem{zhou2019unet++}
Zongwei Zhou, Md~Mahfuzur~Rahman Siddiquee, Nima Tajbakhsh, and Jianming Liang.
\newblock Unet++: Redesigning skip connections to exploit multiscale features in image segmentation.
\newblock {\em IEEE transactions on medical imaging}, 39(6):1856--1867, 2019.

\bibitem{zhuang2019evaluation}
Xiahai Zhuang and et~al. Li, Lei.
\newblock Evaluation of algorithms for multi-modality whole heart segmentation: an open-access grand challenge.
\newblock {\em Medical Image Analysis}, 58:101537, 2019.

\bibitem{zou2023segment}
Xueyan Zou, Jianwei Yang, Hao Zhang, Feng Li, Linjie Li, Jianfeng Gao, and Yong~Jae Lee.
\newblock Segment everything everywhere all at once.
\newblock {\em arXiv preprint arXiv:2304.06718}, 2023.

\end{thebibliography}
}

\end{document}